\documentclass[twoside]{article}

\usepackage[preprint]{aistats2027}

\usepackage[round]{natbib}

\usepackage[utf8]{inputenc}
\DeclareUnicodeCharacter{2192}{\ensuremath{\rightarrow}}
\usepackage[T1]{fontenc}
\usepackage{amsmath,amssymb}
\usepackage{booktabs}
\usepackage{multirow}
\usepackage{longtable}
\usepackage{array}
\usepackage{url}
\usepackage{xurl}
\usepackage{graphicx}
\usepackage{xcolor}
\usepackage{pifont}
\usepackage{upquote}  %
\usepackage{tikz}
\usepackage{placeins}
\usepackage{hyperref}
\newcommand{\paperAuthors}{Shenqi Zhou\textsuperscript{1,2,*},\quad Qiran Zou\textsuperscript{1,*},\quad Wenhao Zhao\textsuperscript{1},\quad Zhenyu Liao\textsuperscript{3},\quad Dianbo Liu\textsuperscript{1,\dag}}
\newcommand{\paperAffiliations}{\textsuperscript{1}National University of Singapore\quad \textsuperscript{2}Zhejiang University\quad \textsuperscript{3}Amazon\\[2pt]
{\small\textsuperscript{*}Equal contribution\quad \textsuperscript{\dag}Corresponding author\quad Contact: \href{mailto:shenqi_zhou@zju.edu.cn}{shenqi\_zhou@zju.edu.cn}, \href{mailto:dianbo@nus.edu.sg}{dianbo@nus.edu.sg}}}
\newcommand{\paperAuthorMetadata}{Shenqi Zhou, Qiran Zou, Wenhao Zhao, Zhenyu Liao, Dianbo Liu}
\newcommand{\paperRunningAuthors}{Zhou et al.}

\hypersetup{
  hidelinks,
  pdftitle={Auto-MD-bench: Evaluating Autoresearch Agents for Real-World Medical Tasks},
  pdfauthor={\paperAuthorMetadata}
}

\newcommand{\bench}{Auto-MD-bench}

\DeclareRobustCommand{\cmark}{\ding{51}}
\DeclareRobustCommand{\xmark}{\textcolor{black!40}{\ding{55}}}
\newsavebox{\pmarkbox}
\AtBeginDocument{\sbox{\pmarkbox}{\tikz[baseline=-0.55ex]{\draw[line width=0.5pt] (0,0) circle (0.6ex);\fill (0,0.6ex) arc (90:270:0.6ex) -- cycle;}}}

\AddToHook{env/table/begin}{\setlength{\belowcaptionskip}{6pt}}
\AddToHook{env/table*/begin}{\setlength{\belowcaptionskip}{6pt}}

\runningtitle{Auto-MD-bench}
\runningauthor{\paperRunningAuthors}
\begin{document}

\twocolumn[
\aistatstitle{\bench: Evaluating Autoresearch Agents for Real-World Medical Tasks}
\aistatsauthor{\paperAuthors}
\aistatsaddress{\paperAffiliations}
]

\begin{abstract}
Autonomous agents are increasingly capable of developing machine learning (ML) models end to end.
In medical practice, however, strong performance on the target metric does not fully characterize how good the models are in real-world use cases. For example, medical ML models should also provide well-calibrated risk estimates and remain robust across patient populations. 
Yet autonomous model development is typically guided by task objectives that emphasize simple predictive discrimination (e.g. AUROC). 
This raises a key question: 
are these objectives sufficient to guide agents toward models that meet the broader requirements of real-world medical use?
In this study, we introduce \bench{}, a benchmark of 15 real-world medical tasks from 10 clinical datasets, spanning survival analysis,
classification and sequential prediction, with external validation across cohorts, hospitals and time periods.
We evaluate four agents: AIDE, AI Scientist v2, Codex, and Autoresearch.
The strongest agents achieve competitive primary scores, yet their predicted risks are less calibrated than those of the simple baseline models, which could lead to severe medical accidents. Moreover, models with nearly identical primary scores can exhibit markedly different calibration, while performance varies substantially across population shifts. 
These findings show that primary task performance alone is insufficient to assess autonomous clinical modeling and multiple different metrics widely used in healthcare practice need to be taken into consideration.
\end{abstract}

\section{Introduction}
\label{sec:intro}

Autonomous research agents can increasingly develop machine learning models end to end, from data inspection and model selection to implementation, validation, and iterative refinement \citep{chan2025mle,miller2025bioml}.
Such capabilities are particularly consequential in medicine, where prediction models use patient data to estimate clinical outcomes and inform patient care \citep{steyerberg2009clinical}.
Automating their development could reduce the effort required to build models for different outcomes and patient populations, but it also delegates clinically consequential modeling decisions to autonomous agents.\footnote{Code: {\urlstyle{same}\url{https://github.com/Edwin0080/Auto-MD-bench}}. Data for 10 of the 15 tasks: {\urlstyle{same}\url{https://huggingface.co/datasets/Edwin0080/Auto-MD-bench}}.}

In real-world medical settings, however, strong performance on a target metric does not fully characterize a model's suitability for medical use.
A model may distinguish patients at higher and lower risk while providing inaccurate estimates of their absolute risk, potentially misleading screening or treatment decisions \citep{van2019calibration}.
Its performance may also change when applied to patients from another cohort, hospital, or time period \citep{royston2013external}.
These aspects of model quality are not necessarily captured by the primary task objective that typically guides autonomous model development.
This raises a key question: \emph{are standard task objectives sufficient to guide agents toward models suitable for real-world medical use?}

Existing benchmarks evaluate autonomous model development across general
machine learning and biomedical tasks, but largely focus on primary task
performance. A unified evaluation of primary performance, risk calibration,
and external validation is still missing (Table~\ref{tab:compare}). This leaves
open whether models optimized for strong task performance also satisfy broader
requirements for medical use. We address this gap by evaluating these
properties together and analyzing how agents' modeling choices relate to the
resulting differences.

\begin{figure*}[t]
\centering
\includegraphics[width=\textwidth]{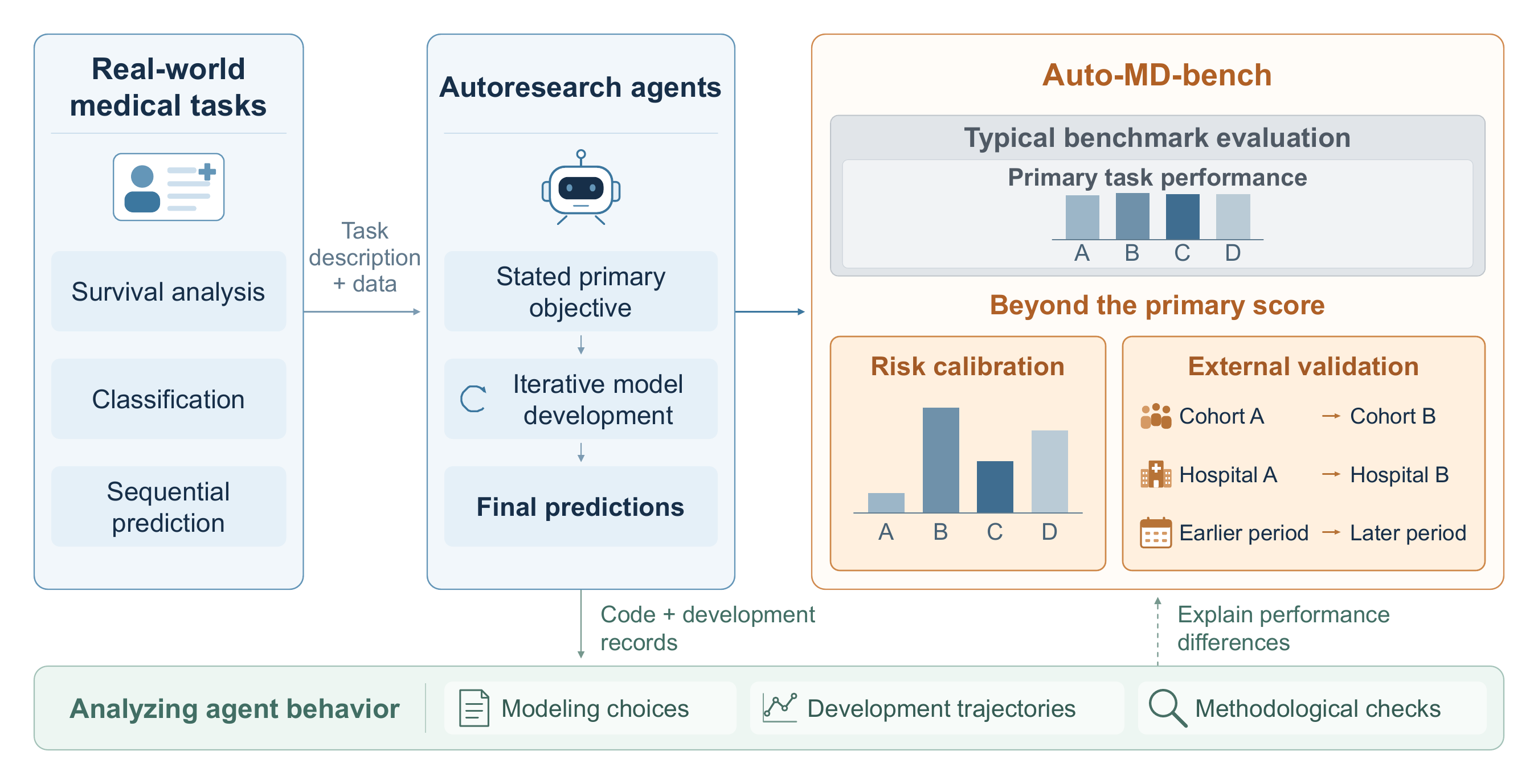}
\caption{Overview of \bench{}.
Agents develop medical prediction models under a stated primary objective.
We evaluate primary task performance, risk calibration, and performance under population shift, and analyze code and development records to help explain performance differences.
External validation uses test data from a different cohort, hospital, or time period; the corresponding setting without population shift uses training and test data from the same population.}
\label{fig:overview}
\end{figure*}

We introduce \bench{}, comprising 15 medical prediction tasks from 10 clinical
datasets spanning survival analysis, classification, and sequential prediction,
with external validation across cohorts, hospitals, and time periods. Agents
choose their modeling and validation procedures using labeled training data,
while test outcomes remain hidden throughout model development. We evaluate the
resulting models on primary task performance, risk calibration, and performance
under population shift, and analyze their code and development records to
characterize how the agents build and revise their models.
Figure~\ref{fig:overview} summarizes the benchmark workflow.

We evaluate four agents: AIDE, AI Scientist v2, Codex, and Autoresearch.
Both Codex and Autoresearch exceed the simple baseline models in mean primary
score, while their mean calibration errors are roughly three times as large.
Across agents, models with similar primary scores can exhibit markedly different
calibration, and performance varies substantially across population shifts.
Analysis of the developed models identifies modeling practices associated with
these calibration differences. In an additional experiment on UCI Diabetes,
explicitly including calibration in the stated objective changes agents'
modeling choices and reduces mean ECE from 0.172 to 0.016 while maintaining
similar AUROC. Together, these results show that the primary task objective
provides an incomplete specification of model quality for medical use.

Our contributions are:
\begin{itemize}

\item We introduce \bench{}, a benchmark of 15 medical prediction tasks from
10 clinical datasets that evaluates autonomous model development beyond the
primary task metric, including risk calibration and external validation across
cohorts, hospitals, and time periods.

\item Across four agents, we show that strong primary performance can mask
substantial differences in calibration and performance under population shift.
We further show in a controlled experiment that explicitly incorporating
calibration into the objective can substantially improve calibration while
maintaining similar predictive discrimination.

\item We extend benchmark evaluation beyond final model outputs by capturing
agents' code and development records, enabling systematic analysis of how
autonomous agents construct, validate, and revise medical prediction models.

\end{itemize}

\section{Related Work}
\label{sec:related}

\paragraph{Autonomous Model Development.}
Benchmarks for autonomous model development evaluate agents through executable machine learning tasks and the performance of their resulting solutions.
MLAgentBench measures agents' ability to improve existing implementations, while MLE-bench evaluates end-to-end model development through Kaggle competitions \citep{huang2023mlagentbench,chan2025mle}.
DSBench covers both data analysis and predictive modeling \citep{jing2025dsbench}.
Beyond final task performance, FML-bench compares research strategies within a common experimental framework and analyzes how agents explore and select candidate solutions \citep{zou2026fml}.

\paragraph{Automated Medical Model Development.}
Automated medical model development spans conventional AutoML pipelines and LLM-driven systems that generate and revise modeling code.
AutoPrognosis 2.0 automates preprocessing, model selection, and hyperparameter tuning for diagnostic and prognostic modeling \citep{imrie2023autoprognosis}.
More recent work has used LLMs to develop clinical classifiers from natural language instructions \citep{tayebi2024large}, construct interpretable clinical scores \citep{estevez2026automatic}, and revise cardiovascular prediction models with evaluation on external cohorts \citep{dhingra2026autonomous}.

\paragraph{Biomedical Agent Benchmarks.}
Biomedical agent benchmarks cover activities ranging from model development to coding-intensive data analysis and broader healthcare workflows.
BioML-bench and BioXArena evaluate model development across diverse biomedical domains, including patient-level prediction \citep{miller2025bioml,li2026bioxarena}.
MedAgentGym emphasizes code-centric biomedical data science, whereas HealthAgentBench and MedAgentBoard evaluate agents across broader healthcare workflows \citep{xu2026medagentgym,liu2026healthagentbench,zhu2026medagentboard}.
AutoHealth and ACP-Bench report calibration analyses alongside predictive performance \citep{xia2026autohealth,lin2026evaluating}, while BioXArena includes evaluation across imaging centers \citep{li2026bioxarena}.
Table~\ref{tab:compare} summarizes calibration and external validation in these benchmarks.

\paragraph{Evaluation for Medical Use.}
Clinical prediction research distinguishes discrimination from risk calibration \citep{steyerberg2009clinical,van2019calibration}.
External validation assesses performance across patient populations \citep{royston2013external}.
These perspectives motivate evaluating agent-developed medical models beyond primary task scores and relating their performance to modeling practices.

\begin{table*}[t]
\centering
\caption{Coverage of evaluation dimensions relevant to medical use in related agent benchmarks.
\cmark: reported; \xmark: not reported. Calibration denotes explicit assessment of probability calibration; external validation uses a distinct cohort, site, or time period. $^{\dagger}$ MedAgentGym reports task categories.}
\label{tab:compare}
\footnotesize
\setlength{\tabcolsep}{5.5pt}
\begin{tabular}{@{}llrcc@{}}
\toprule
 & & & \multicolumn{2}{c}{Evaluation for medical use} \\
\cmidrule(l){4-5}
Benchmark & Domain & Tasks & Calibration & External validation \\
\midrule
MLE-bench & General ML (Kaggle) & 75 & \xmark & \xmark \\
HealthAgentBench & Healthcare workflows & 54 & \xmark & \xmark \\
MedAgentBoard & Diverse medical tasks & 100 & \xmark & \xmark \\
MedAgentGym & Biomedical data science & 129$^{\dagger}$ & \xmark & \xmark \\
AutoHealth & Health prediction & 17 & \cmark & \xmark \\
ACP-Bench & Computational pathology & 41 & \cmark & \xmark \\
BioML-bench & Biomedical ML & 24 & \xmark & \xmark \\
BioXArena & Biomedical ML & 76 & \xmark & \cmark \\
\midrule
\textbf{\bench{} (ours)} & Clinical prediction & 15 & \cmark & \cmark \\
\bottomrule
\end{tabular}
\end{table*}

\section{The Benchmark}
\label{sec:benchmark}

\bench{} evaluates autoresearch agents that develop models for medical prediction tasks with a specified primary metric.
To test whether the primary metric adequately specifies model quality for medical use, we assess primary task performance alongside risk calibration and performance under population shift.
We also examine the agents' code and development records to characterize their modeling choices and relate them to calibration.
Figure~\ref{fig:overview} shows the workflow from task inputs through autonomous model development to evaluation.

\subsection{Medical Tasks for Autoresearch Agents}
\label{sec:tasks}
The benchmark comprises 15 medical prediction tasks from 10 clinical datasets, using structured patient data to predict clinical outcomes.
The tasks cover survival analysis, classification and sequential prediction.
Agents provide survival curves for survival analysis, event probabilities for classification, and hourly probabilities and alarms for sequential prediction.
The full task list, data sources, outcomes, sample sizes, and primary metrics are reported in Appendix~\ref{app:tasks} (Table~\ref{tab:tasks}).

Ten tasks use internal validation on test data from the source population.
Five use external validation on data from another cohort, hospital or period.
One external validation task also provides 500 labeled patients from the target hospital for model development.

\subsection{Medical Task Workflow}
\label{sec:env}

Each task provides a description, labeled training data, test features, null predictions in the required submission format, and a format checker.
The description specifies the prediction target, primary performance metric and task rules.
Agents choose their modeling approach, write and execute code, train and validate models using the training data, and revise their models using execution results and validation scores.
Test labels and test scores are unavailable throughout model development.

Agents submit prediction files for evaluation \citep{chan2025mle}.
For each run, we select one submission using the agent-specific rules in Table~\ref{tab:agents}.
After model development is complete, we score the selected predictions outside the agent's development environment.
A run is valid if its selected submission passes the submission format checks; missing, unreadable or malformed files make it invalid.
Appendix~\ref{app:pertask} gives the detailed checks.

\subsection{Evaluation Metrics}
\label{sec:scoring}

\paragraph{Predictive Performance.}
We evaluate survival predictions with adjusted Antolini time-dependent concordance $C^{td}$ \citep{antolini2005time}.
Classification tasks use AUROC, except for PhysioNet 2012, which uses the minimum of sensitivity and positive predictive value \citep{silva2012predicting}.
Sequential prediction tasks use the PhysioNet/CinC 2019 normalized utility, which rewards timely alarms and penalizes false and missed alarms \citep{reyna2019early}.

We normalize each task's raw score $m$ relative to the score $m_0$ of its null predictions:
\begin{equation}
s=\min\!\left(1,\max\!\left(0,\frac{m-m_0}{1-m_0}\right)\right).
\label{eq:norm}
\end{equation}
Here, $m_0=0.5$ for $C^{td}$ and AUROC, and $m_0=0$ for the other metrics.
Penalized scores assign zero to invalid runs; non-penalized scores use valid runs only.
Both weight tasks equally (Table~\ref{tab:main}).
See Appendix~\ref{app:pertask} for details.

\paragraph{Simple Baseline Models.}
We use Cox neural networks implemented in \texttt{pycox} as simple baseline models for survival analysis  \citep{katzman2018deepsurv,kvamme2019time} and \texttt{scikit-learn}'s histogram-based gradient-boosted trees for classification and sequential prediction.
Each simple baseline model uses fixed hyperparameters and the task's training data, and is evaluated with the same metrics and evaluation code as the agent-developed models.
We train each simple baseline model five times per task and average its predictive scores and calibration errors across these repetitions.
Appendix~\ref{app:references} gives the model configurations and training procedures.

\paragraph{Risk Calibration.}
We use expected calibration error (ECE) to compare predicted probabilities with observed outcome frequencies (Table~\ref{tab:ece}).
For survival analysis, we group patients by predicted survival probability at each of three time points.
We compare the mean prediction in each group with its Kaplan--Meier survival estimate, which accounts for censoring.
For sequential prediction, ECE is computed over patient-hours.
Appendix~\ref{app:behavior} defines the bins, time points and aggregation.

\paragraph{External Validation.}
We report normalized scores on the five external validation tasks and their ratios to the corresponding scores without the population shift (Table~\ref{tab:xsite}).
Agents develop models separately for the internal and external validation tasks.
We compare these ratios with those of the simple baseline models.
Appendix~\ref{app:pertask} gives the calculation.

\paragraph{Subgroup Performance.}
We compare performance across age, sex, and recorded race/ethnicity groups, using AUROC for classification and time-dependent concordance for survival analysis.
Appendix~\ref{app:behavior} gives the group definitions and calculations.

\paragraph{Modeling Methodology.}
For valid submissions, we examine the prediction code and supporting development records to identify model families, validation strategies, and approaches to class imbalance and censoring.
We use Claude Opus 5.5 for annotation, with each annotation citing supporting code by file and line number.
We assess annotation consistency through repeated annotation (Table~\ref{tab:app-reading}).
We examine how using averaged ranks as probabilities, class imbalance handling and recalibration relate to differences in ECE from the simple baseline models (Figure~\ref{fig:calib-share}).
Table~\ref{tab:methods} summarizes the modeling methods, and Appendix~\ref{app:reading} defines the calibration groups.
We also compare the code for sampled pairs of submission versions to characterize modeling changes during agent search (Figure~\ref{fig:changes}; Appendix~\ref{app:search}).

\paragraph{Real-Time Rule and Masking Test.}
\label{sec:masking}
At hour $t$, sequential predictions may use only the patient's observations through $t$ and information learned from the training data.
To detect violations of this rule, we mask later observations by removing the end of each patient's record, rerun the code that produced the selected submissions, and compare alarms at the retained hours.
We apply this test to valid submissions from AIDE and AI Scientist v2 on the three sequential prediction tasks.
Appendix~\ref{app:masking} describes the procedure and reports the results.

\section{Experiments}
\label{sec:experiments}
\label{sec:results}

\subsection{Experimental Setup}
\label{sec:setup}

\paragraph{Autoresearch Agents and Harness.}
We evaluate four autoresearch agents: Codex, Autoresearch, AIDE, and AI Scientist v2.
Codex is a general-purpose coding agent \citep{openai_codex}.
Autoresearch uses Codex with the autoresearch protocol \citep{karpathy2026autoresearch}, which instructs Codex to build an initial model, then propose and evaluate one change at a time, retaining changes that improve the primary metric on its validation data.
AIDE uses greedy tree search \citep{jiang2025aide}, whereas AI Scientist v2 uses best-first tree search organized into four stages: initial implementation, tuning, experimentation, and ablation \citep{yamada2025ai}.

\paragraph{Experimental Conditions.}
In the main experiments, all four agents use Qwen3.8 Flash \citep{qwen38flash}.
We evaluate each agent three times on each of the 15 tasks, with a 12-hour wall-clock budget for each repetition.
Each repetition uses a dedicated virtual machine with eight CPU threads and 16 GB of RAM.
All agents receive the same task description and data for a given task.
They run in Apptainer containers with the same preinstalled software and no internet access.

\paragraph{Implementation Details.}
We adapt each framework's input interface to provide the benchmark's task descriptions and data.
For AI Scientist v2, we use its experiment stages and disable paper writing and review.
To apply the 12-hour budget, we raise AIDE's iteration limit and allocate time across AI Scientist v2's stages in proportion to their original iteration quotas.
Appendix~\ref{app:agents} provides configuration and implementation details, including software versions, a task description example, and the autoresearch instructions.

\subsection{Performance on Medical Tasks}
\label{sec:overall}
\begin{table*}[t]
\centering
\caption{Primary task performance on 15 tasks with Qwen3.8 Flash. Scores and valid run rates are mean $\pm$ SD over three repetitions; ranks are mean $\pm$ SD across tasks. Penalized scores assign zero to invalid runs; non-penalized scores exclude them. Bold: best agent value per metric.}
\label{tab:main}
\footnotesize
\setlength{\tabcolsep}{6pt}
\begin{tabular}{@{}lccccc@{}}
\toprule
 & \multicolumn{2}{c}{Normalized score $\uparrow$} & \multicolumn{2}{c}{Mean rank $\downarrow$} & Valid \\
\cmidrule(lr){2-3}\cmidrule(lr){4-5}
System & penalized & non-pen. & penalized & non-pen. & runs \\
\midrule
\textit{Simple baseline models} & \multicolumn{2}{c}{0.487} & \multicolumn{2}{c}{--} & -- \\
\midrule
AIDE & 0.474 $\pm$ 0.032 & 0.474 $\pm$ 0.032 & 2.93 $\pm$ 0.88 & 3.13 $\pm$ 0.99 & 100.0\% $\pm$ 0.0\% \\
Codex & 0.517 $\pm$ 0.003 & 0.517 $\pm$ 0.003 & \textbf{1.60} $\pm$ 0.51 & \textbf{1.60} $\pm$ 0.51 & 100.0\% $\pm$ 0.0\% \\
Autoresearch & \textbf{0.518} $\pm$ 0.006 & \textbf{0.518} $\pm$ 0.006 & 1.80 $\pm$ 0.77 & 1.80 $\pm$ 0.77 & 100.0\% $\pm$ 0.0\% \\
AI Scientist v2 & 0.355 $\pm$ 0.064 & 0.467 & 3.67 $\pm$ 0.82 & 3.47 $\pm$ 0.83 & 73.3\% $\pm$ 13.3\% \\
\bottomrule
\end{tabular}
\end{table*}

The general coding agents achieve higher mean normalized scores than the research agents (Table~\ref{tab:main}).
Codex and Autoresearch score 0.517 and 0.518, respectively, compared with 0.474 for AIDE and 0.355 for AI Scientist v2.
Codex and Autoresearch exceed the mean score of the simple baseline models (0.487), and each exceeds the simple baseline score on 14 of the 15 tasks (Appendix~\ref{app:pertask}).
Codex and Autoresearch have similar mean scores, but Codex achieves a better mean rank across tasks (1.60 versus 1.80).

AI Scientist v2 is the only agent to produce invalid submissions, with 12 invalid submissions across 45 experiments.
Excluding these submissions raises its mean normalized score from 0.355 to 0.467, close to AIDE's 0.474 but below Codex and Autoresearch.

\subsection{Risk Calibration}
\label{sec:calibration}
\begin{table*}[!htbp]
\centering
\caption{Expected calibration error (ECE; lower is better) by agent and task class. Bold: lowest value per column.}
\label{tab:ece}
\footnotesize
\setlength{\tabcolsep}{8pt}
\begin{tabular}{@{}lcccc@{}}
\toprule
& All & Survival analysis & Classification & Sequential prediction \\
\midrule
AIDE & 0.128 & 0.045 & 0.280 & 0.066 \\
Codex & 0.086 & 0.076 & 0.113 & 0.065 \\
Autoresearch & 0.075 & 0.061 & 0.087 & 0.090 \\
AI Scientist v2 & 0.072 & 0.048 & 0.102 & 0.078 \\
\midrule
Simple baseline models & \textbf{0.027} & \textbf{0.027} & \textbf{0.031} & \textbf{0.018} \\
\bottomrule
\end{tabular}
\end{table*}

\begin{figure*}[!t]
\centering
\includegraphics[width=\textwidth]{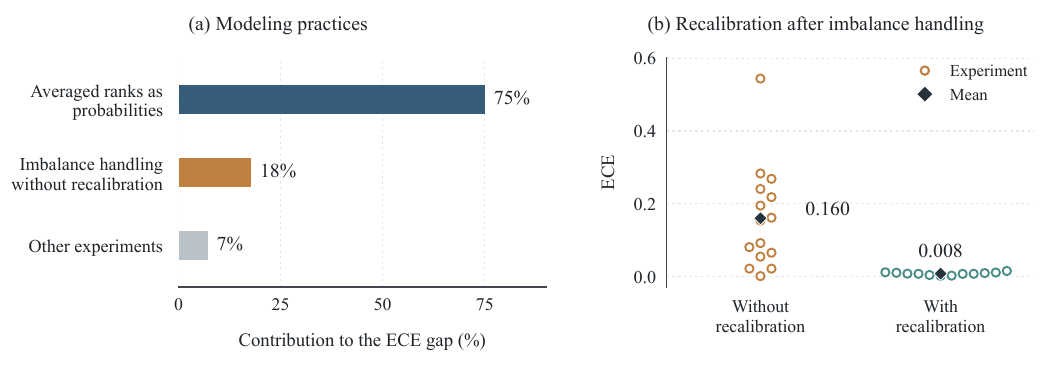}
\caption{Calibration error and modeling practices. (a) Contributions to the ECE gap relative to task-specific simple baseline models. (b) ECE after agents handle class imbalance, with or without recalibration.
Circles show individual experiments; diamonds show group means.
Group definitions are in Appendix~\ref{app:reading}.}
\label{fig:calib-share}
\end{figure*}

High primary scores do not imply well-calibrated predictions.
Codex and Autoresearch exceed the simple baseline models in mean primary score, but have roughly three times their average ECE (0.086 and 0.075 versus 0.027; Table~\ref{tab:ece}).
For example, two individual UCI Diabetes submissions from Codex and AI Scientist v2 have nearly identical AUROC values of 0.670 and 0.671, but ECE values of 0.006 and 0.154, respectively.

Most of the ECE difference from the simple baseline models is concentrated in two groups of experiments (Figure~\ref{fig:calib-share}).
Experiments using averaged ranks as probabilities account for 75\% of this difference and 14\% of valid experiments.
Among the remaining experiments, those that handle class imbalance without recalibration account for a further 18\% of the difference and 9\% of all valid experiments.
Averaged ranks express relative ordering rather than event probabilities, while handling class imbalance changes the relative weight of events and non-events during training and can distort the resulting probabilities.
For experiments that handle class imbalance but do not use averaged ranks, mean ECE is 0.008 with subsequent recalibration and 0.160 without it.

\begin{table}[!htbp]
\centering
\caption{Calibration under three evaluation objectives on UCI Diabetes (mean $\pm$ SD over valid runs). Bold: best value across objectives.}
\label{tab:calib-control}
\footnotesize
\setlength{\tabcolsep}{4pt}
\begin{tabular}{@{}lcc@{}}
\toprule
Objective & AUROC $\uparrow$ & ECE $\downarrow$ \\
\midrule
AUROC & 0.676 $\pm$ 0.023 & 0.172 $\pm$ 0.211 \\
Brier score & 0.640 $\pm$ 0.107 & 0.061 $\pm$ 0.157 \\
AUROC and ECE & \textbf{0.683} $\pm$ 0.016 & \textbf{0.016} $\pm$ 0.023 \\
\bottomrule
\end{tabular}
\end{table}

\begin{table*}[!htbp]
\centering
\caption{Performance under population shift. Entries are mean normalized scores; parentheses give ratios to the corresponding score without the population shift. Bold: best score per setting; $^{*}$ one valid run.}
\label{tab:xsite}
\footnotesize \renewcommand{\arraystretch}{1.1}
\setlength{\tabcolsep}{3pt}
\begin{tabular}{@{}>{\raggedright\arraybackslash}p{0.26\textwidth}ccccc@{}}
\toprule
External validation setting & \shortstack{Simple baseline\\models} & AIDE & Codex & Autoresearch & AI Scientist v2 \\
\midrule
Rotterdam $\to$ GBSG (cohort) & 0.372 (100\%) & 0.306 (86\%) & \textbf{0.385} (99\%) & 0.371 (97\%) & 0.332 (85\%) \\
SEER (temporal) & 0.724 (99\%) & 0.730 (97\%) & \textbf{0.745} (103\%) & 0.735 (99\%) & 0.670$^{*}$ (--) \\
eICU (hospital) & 0.736 (99\%) & 0.748 (99\%) & \textbf{0.751} (100\%) & \textbf{0.751} (100\%) & 0.705 (100\%) \\
Sepsis (hospital) & 0.199 (52\%) & 0.168 (41\%) & 0.231 (52\%) & \textbf{0.268} (61\%) & 0.228$^{*}$ (63\%) \\
Sepsis + 500 target labels & 0.248 (65\%) & 0.217 (54\%) & \textbf{0.299} (67\%) & 0.253 (57\%) & 0.124$^{*}$ (34\%) \\
\bottomrule
\end{tabular}
\end{table*}

To test whether explicitly specifying calibration changes model development, we compare the main experiment's AUROC objective on UCI Diabetes with two alternatives: Brier score, which reflects both discrimination and calibration, and AUROC with ECE.
We specify each objective in the task description and use three repetitions of each of the four agents, with the same LLM, data, and time budget.
Adding ECE to the stated AUROC objective reduces mean ECE from 0.172 to 0.016 while maintaining similar mean AUROC (0.676 versus 0.683; Table~\ref{tab:calib-control}).
In this condition, neither using ranks as probabilities nor handling class imbalance without recalibration is observed.
With Brier score, mean ECE also decreases, to 0.061, but mean AUROC is lower at 0.640.
These results show that including calibration in the stated objective can change modeling choices and improve calibration on this task.
Appendix~\ref{app:calib-control} provides results by agent and code annotations.

\subsection{Generalization across Different Populations}
\label{sec:shift}

\paragraph{External Validation.}
Performance under population shift varies substantially across tasks (Table~\ref{tab:xsite}).
On Rotterdam $\to$ GBSG, scores under population shift are 85--86\% of the corresponding scores without the shift for the research agents and 97--99\% for the coding agents.
Ratios are lower on both sepsis tasks: 41--63\% on \emph{Sepsis, new hospital} and 34--67\% on \emph{Sepsis, 500 local labels}.
On \emph{SEER, later period} and \emph{eICU, new hospitals}, performance changes little under the population shift, with ratios close to 100\%.

Codex and Autoresearch achieve higher scores under population shift than the two research agents on all five tasks.
Codex exceeds the simple baseline score on every task, while Autoresearch exceeds it on four and closely matches it on Rotterdam $\to$ GBSG.
The corresponding ratios for Codex and Autoresearch are also close to those of the simple baseline models.
The ratios for AIDE and AI Scientist v2 are approximately 8 and 9 percentage points lower than the corresponding simple baseline ratios, respectively, on average.
These results show that performance under population shift is an additional property that must be evaluated rather than inferred from the primary score alone.

\paragraph{Subgroup Performance.}
Aggregate scores also mask differences between patient subgroups.
In the four ICU mortality tasks, the oldest age tertile is the lowest-scoring age group in every valid experiment (Appendix Table~\ref{tab:app-subgroups}).
The mean difference between the highest- and lowest-scoring age groups ranges from 0.065 to 0.084, similar to the 0.065--0.078 differences for the simple baseline models.
Across all tasks, agents check performance by age or sex in only six of 168 valid experiments.
Appendix~\ref{app:behavior} reports the full subgroup results, including sex and race/ethnicity.

\subsection{Machine Learning Model Choice by Different Autoresearch Agents}
\label{sec:search}
\begin{table*}[t]
\centering
\caption{Modeling methods in valid experiments (mean $\pm$ SD over three repetitions; categories can overlap).}
\label{tab:methods}
\footnotesize
\setlength{\tabcolsep}{6pt}
\begin{tabular}{@{}lcccc@{}}
\toprule
Method & AIDE & Codex & Autoresearch & AI Scientist v2 \\
\midrule
\multicolumn{5}{@{}l}{\emph{Model family}} \\
Gradient-boosted trees & 87\% $\pm$ 0\% & 89\% $\pm$ 8\% & 89\% $\pm$ 8\% & 76\% $\pm$ 13\% \\
Neural networks & 11\% $\pm$ 8\% & 31\% $\pm$ 17\% & 38\% $\pm$ 10\% & 36\% $\pm$ 10\% \\
\addlinespace[2pt]
\multicolumn{5}{@{}l}{\emph{Ensembling and validation}} \\
Averaging across seeds or folds & 62\% $\pm$ 14\% & 89\% $\pm$ 4\% & 89\% $\pm$ 10\% & 31\% $\pm$ 7\% \\
Single model only & 13\% $\pm$ 12\% & 2\% $\pm$ 4\% & 4\% $\pm$ 8\% & 36\% $\pm$ 10\% \\
Single holdout split only & 9\% $\pm$ 4\% & 4\% $\pm$ 4\% & 13\% $\pm$ 7\% & 37\% $\pm$ 16\% \\
\addlinespace[2pt]
\multicolumn{5}{@{}l}{\emph{Calibration}} \\
Probability calibration & 2\% $\pm$ 4\% & 22\% $\pm$ 10\% & 22\% $\pm$ 8\% & 14\% $\pm$ 12\% \\
\bottomrule
\end{tabular}
\end{table*}

\begin{figure*}[!t]
\centering
\includegraphics[width=\textwidth]{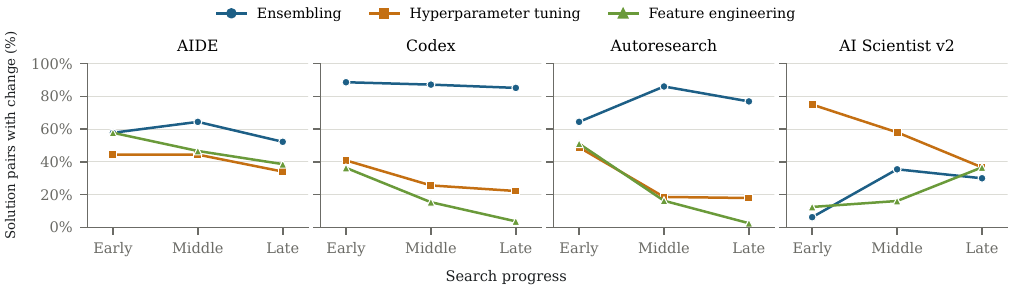}
\caption{Modeling changes during agent search.
Points show the percentage of sampled solution pairs containing each type of modeling change.
Early, middle, and late indicate relative positions in the chronologically ordered sequence of solution pairs within each run, rather than equal intervals of elapsed time.
Pairs may contain multiple types of change.}
\label{fig:changes}
\end{figure*}

We examine the methods used to build these models to assess their handling of validation, probability calibration, and clinical information.

\paragraph{Model Families and Ensembles.}
Gradient-boosted trees dominate model choice, appearing in 76--89\% of valid experiments across the four agents (Table~\ref{tab:methods}).
The agents differ more in how they combine models: Codex and Autoresearch average predictions across seeds or folds in 89\% of experiments, compared with 62\% for AIDE and 31\% for AI Scientist v2.
AI Scientist v2 uses a single model in 36\% of experiments, compared with 2--13\% for the other agents.

\paragraph{Validation and Calibration.}
AI Scientist v2 also relies on a single holdout split more often: 37\% of experiments, versus 4--13\% for the other agents.
Probability calibration is uncommon across all four agents, appearing in 2--22\% of experiments.

\paragraph{Modeling Changes during Agent Search.}
Changes to ensembling are frequent for Codex and Autoresearch, appearing in 64--89\% of sampled pairs (Figure~\ref{fig:changes}).
Hyperparameter tuning is less frequent in late than in early pairs for all four agents.
Feature changes also decline in Codex and Autoresearch, reaching 3--4\% in late pairs.
AI Scientist v2, which follows a predefined four-stage workflow, instead shows more feature changes in later pairs.

\paragraph{Censoring and Temporal Information.}
Code review identifies errors in handling censored outcomes in eight of the 79 valid survival submissions.
The errors occur in ensemble members and include treating censored patients as observed deaths or as known survivors at later time points, and using incorrect censoring weights.
All eight submissions exceed their task's simple baseline score despite these errors.
Thus, success on the primary score can coexist with errors in how the underlying medical prediction problem is modeled.

The masking test detects violations of the real-time rule in three of the 13 valid sequential prediction experiments tested for AIDE and AI Scientist v2.
All three occur in AIDE and use information from later hours to determine earlier predictions or alarms.
Appendix~\ref{app:masking} reports the cases and test results.

\subsection{Effect of the LLM}
\label{sec:llms}
\begin{table}[!htbp]
\centering
\caption{Codex with three LLMs. Scores are mean $\pm$ SD over three repetitions; ECE is averaged across tasks. Bold: best value per metric.}
\label{tab:llm}
\footnotesize
\setlength{\tabcolsep}{5pt}
\begin{tabular}{@{}lcc@{}}
\toprule
LLM & Normalized score $\uparrow$ & ECE $\downarrow$ \\
\midrule
Qwen3.8 Flash & 0.517 $\pm$ 0.003 & 0.086 \\
gpt-6-luna & 0.517 $\pm$ 0.004 & 0.041 \\
Gemini 3.8 Flash & \textbf{0.520} $\pm$ 0.006 & \textbf{0.029} \\
\bottomrule
\end{tabular}
\end{table}

Keeping Codex fixed, we evaluate two additional LLMs on all 15 tasks, with three repetitions per task, the same data, and a 12-hour time budget.
Mean normalized scores are similar across Qwen3.8 Flash, gpt-6-luna, and Gemini 3.8 Flash, ranging from 0.517 to 0.520 (Table~\ref{tab:llm}).
Calibration differs substantially: mean ECE is 0.086, 0.041, and 0.029, respectively.
Gemini 3.8 Flash's mean ECE is close to the simple baseline models' mean ECE of 0.027.
The three LLMs achieve nearly identical mean primary scores, but their calibration differs substantially.
Appendix~\ref{app:pending} reports submission validity and token usage.

\section{Conclusion}
\label{sec:conclusion}

We introduced \bench{} to examine whether the common practice of guiding
autonomous agents with a single primary metric is sufficient for developing
models that satisfy broader requirements for medical use. Across 15 tasks from
10 clinical datasets, models with similar primary performance could differ
markedly in risk calibration and in performance under population shift. Our
analysis connected these differences to concrete modeling choices, while
explicitly incorporating calibration into model development substantially
improved calibration while maintaining similar discrimination. These results
show that primary task performance provides only a partial view of the models
produced by autonomous agents. More broadly, the metric used to guide model
development can influence which properties of the resulting model are
prioritized. Evaluating autonomous medical model development should therefore
look beyond a single primary metric and consider the broader properties required
by the intended medical use.

\label{main:end}
\clearpage
\subsection*{AI Use Statement}
Claude and OpenAI Codex assisted with experiment design, implementation,
data preparation, literature searches, analysis of experimental results,
figure preparation, and manuscript drafting and revision. ChatGPT assisted with manuscript revision. Claude Opus 5.5
annotated the agents' code using the procedure in Appendix~\ref{app:reading}.
The LLMs evaluated in the benchmark are described in
Section~\ref{sec:setup} and Appendix~\ref{app:pending}.
The authors take responsibility for the final content of this work.

\bibliographystyle{plainnat}
\bibliography{refs}

\clearpage
\appendix
\thispagestyle{empty}
\onecolumn
\aistatstitle{\bench: Supplementary Materials}

\section{Datasets}
\label{app:tasks}
\begin{table}[htbp]
\centering
\caption{The 15 tasks in \bench{}. Sample sizes are patients unless noted; UCI Diabetes counts encounters. Matching symbols indicate shared test sets. $C^{td}$: time-dependent concordance; Utility: PhysioNet/CinC 2019 challenge score.}
\label{tab:tasks}
\footnotesize
\setlength{\tabcolsep}{3pt}
\begin{tabular}{@{}llllll@{}}
\toprule
Task & Data source & Outcome & Validation & Metric & Train / test \\
\midrule
\multicolumn{6}{@{}l}{\emph{Survival analysis}} \\
SUPPORT & Cohort & Death & Internal & $C^{td}$ & 7{,}098 / 1{,}775 \\
METABRIC & Cohort & Death & Internal & $C^{td}$ & 1{,}523 / 381 \\
Rotterdam & Cohort & Recurrence or death & Internal & $C^{td}$ & 2{,}385 / 597 \\
Rotterdam $\to$ GBSG & Cohort $\to$ trial & Recurrence or death & External (cohort) & $C^{td}$ & 2{,}982 / 686 \\
SEER$^{\dagger}$ & Registry & Breast-cancer death & Internal & $C^{td}$ & 232{,}625 / 49{,}291 \\
SEER, later period$^{\dagger}$ & Registry & Breast-cancer death & External (period) & $C^{td}$ & 183{,}334 / 49{,}291 \\
NHANES & Survey & Death & Internal & $C^{td}$ & 11{,}639 / 2{,}910 \\
\midrule
\multicolumn{6}{@{}l}{\emph{Classification}} \\
UCI Diabetes & EHR & 30-day readmission & Internal & AUROC & 81{,}347 / 20{,}419 \\
PhysioNet 2012 & EHR (ICU) & In-hospital death & Internal & $\min(\mathrm{Se},\mathrm{PPV})$ & 4{,}000 / 8{,}000 stays \\
MIMIC-IV & EHR (ICU) & In-hospital death & Internal & AUROC & 41{,}471 / 10{,}368 \\
eICU$^{\ddagger}$ & EHR (ICU) & In-hospital death & Internal & AUROC & 94{,}823 / 11{,}133 stays \\
eICU, new hospitals$^{\ddagger}$ & EHR (ICU) & In-hospital death & External (hospital) & AUROC & 83{,}626 / 11{,}133 stays \\
\midrule
\multicolumn{6}{@{}l}{\emph{Sequential prediction}} \\
Sepsis & EHR (ICU) & Sepsis onset & Internal & Utility & 32{,}268 / 8{,}068 \\
Sepsis, new hospital$^{\S}$ & EHR (ICU) & Sepsis onset & External (hospital) & Utility & 20{,}336 / 4{,}000 \\
Sepsis, 500 local labels$^{\S}$ & EHR (ICU) & Sepsis onset & External (hospital) & Utility & 20{,}336 + 500 / 4{,}000 \\
\bottomrule
\end{tabular}
\end{table}

\begin{table}[htbp]
\centering
\caption{Source datasets.}
\label{tab:app-datasets}
\footnotesize
\setlength{\tabcolsep}{4pt}
\begin{tabular}{@{}>{\raggedright\arraybackslash}p{0.365\textwidth}>{\raggedright\arraybackslash}p{0.47\textwidth}l@{}}
\toprule
Dataset & Content & Access \\
\midrule
SUPPORT  \citep{connors1995controlled}
 & $\sim$9,100 seriously ill inpatients, 5 US hospitals, 1989--94 & public \\
METABRIC  \citep{curtis2012genomic}
 & $\sim$2,000 breast tumors with genomic profiles & public \\
Rotterdam, GBSG  \citep{royston2013external}
 & breast cancer registry (1978--93) and trial (1984--89) & public \\
UCI Diabetes  \citep{strack2014impact,diabetes_130-us_hospitals_for_years_1999-2008_296}
 & 101,766 encounters involving patients with diabetes, 130 US hospitals, 1999--2008 & public \\
PhysioNet/CinC 2012  \citep{silva2012predicting}
 & 12,000 ICU stays, first 48 h & public \\
PhysioNet/CinC 2019  \citep{reyna2019early,PhysioNet-challenge-2019-1.0.0}
 & hourly ICU records labeled for sepsis onset, 40,336 patients, 2 hospital systems & public \\
MIMIC-IV  \citep{johnson2023mimic,PhysioNet-mimiciv-3.1}
 & Beth Israel Deaconess EHR, 2008--22 & credentialed \\
eICU-CRD  \citep{pollard2018eicu,PhysioNet-eicu-crd-2.0}
 & $\sim$200,000 ICU stays, 208 US hospitals, 2014--15 & credentialed \\
SEER  \citep{nci2026seer}
 & US cancer registry, breast cancer 2010--15 & data agreement \\
NHANES  \citep{nchs2022linked}
 & US health survey linked to death records & public \\
\bottomrule
\end{tabular}
\end{table}

\paragraph{Task Construction.}
Test sets are fixed across runs; Table~\ref{tab:tasks} gives the sample sizes.
Unless specified below, we use an 80/20 patient split with random seed 42.

\subsection{Survival Analysis}

\paragraph{SUPPORT and METABRIC.}
These tasks predict time to death from the processed datasets distributed
with \texttt{pycox} \citep{connors1995controlled,curtis2012genomic}.
SUPPORT has 14 clinical covariates and follow-up in days; METABRIC has
four gene-expression markers and five clinical covariates, with follow-up
in months. Both use generic covariate names and unstratified patient splits.

\paragraph{Rotterdam and Rotterdam $\to$ GBSG.}
The Rotterdam and GBSG cohorts are distributed together in the R
\texttt{survival} package. Both tasks predict recurrence or death from
eight harmonized clinical covariates \citep{royston2013external}.
For Rotterdam, follow-up is the minimum of the recorded recurrence and
death times in days; either outcome counts as an event. The internal task splits
Rotterdam patients with stratification by event status. The external task
trains on the full Rotterdam cohort and tests on the independent GBSG
cohort, using its supplied recurrence-free survival time and event status.

\paragraph{SEER and SEER, Later Period.}
The cohort includes women with a first malignant breast-cancer diagnosis
in 2010--2015. Inputs are 22 coded demographic, tumor and treatment
variables \citep{nci2026seer}. The outcome is breast-cancer death,
with follow-up in months; deaths from other causes are censored at death.
We exclude deaths with unknown cause and retain records flagged as having
complete dates and positive follow-up. Both tasks test on the same randomly selected half of patients
diagnosed in 2014--2015. SEER trains on all remaining patients, whereas
\emph{SEER, later period} trains only on diagnoses from 2010--2013.

\paragraph{NHANES.}
The task predicts all-cause mortality after examination from the
2011--2016 survey cycles linked to mortality follow-up through 2019 \citep{nchs2022linked}. We retain linkage-eligible adults aged at least
20 years with positive follow-up and examination weights. The 25 inputs
include clinical and demographic variables, survey weight, stratum,
primary sampling unit and cycle. Follow-up is measured in months, and the
patient split is stratified by event status.

\paragraph{Survival Time Grids.}
The time grid specifies when models must report survival probabilities.
We use the unique quantiles of event times in the training data at 100 evenly
spaced levels from 0 to 1, adding time zero. The two SEER tasks share the
grid from the 2010--2013 training data.

\subsection{Classification}

\paragraph{UCI Diabetes.}
The task predicts readmission within 30 days of a hospital encounter \citep{strack2014impact}. We remove three fields with more than 40\% missing
values and ordinal-encode categorical variables, leaving 44 predictors.
Patients are split with stratification by whether they ever have a positive
readmission label; all encounters of a patient remain together. Patient
identifiers are retained in training data and removed from test features.

\paragraph{PhysioNet 2012.}
The task predicts in-hospital death from irregular observations
over the first 48 hours of an ICU stay \citep{silva2012predicting}. Inputs
are supplied in long format with observation times, variable names and
values. Challenge set A supplies training data; sets B and C together
form the test set.

\paragraph{MIMIC-IV.}
The task predicts in-hospital death for adults whose first ICU stay lasts
at least 24 hours \citep{johnson2023mimic}. Its 73 inputs combine
first-24-hour vital-sign and laboratory summaries with demographics and
additional clinical variables. The patient split is stratified by the
mortality label.

\paragraph{eICU and eICU, New Hospitals.}
Both tasks predict in-hospital death for adult ICU stays of at least
one day with APACHE predictor records \citep{pollard2018eicu}.
Their 80 inputs combine first-24-hour vital-sign and laboratory summaries
with demographics, clinical assessments and admission information.
We randomly select hospitals until they cover at least 20\% of cohort
stays, then use half of their patients for the shared test set.
The eICU task trains on all non-test patients;
\emph{eICU, new hospitals} trains only on the other hospitals. Every stay
of a test patient is excluded from both training sets. Hospital identifiers
are not supplied as predictors.

\subsection{Sequential Prediction}

\paragraph{Hourly Inputs and Target.}
All three tasks use PhysioNet/CinC 2019 records from hospitals A and B \citep{reyna2019early}. Each patient-hour has 40 variables covering vital
signs, laboratory measurements and demographic or administrative
information, plus a hospital indicator. The supplied label becomes
positive six hours before sepsis onset and remains positive thereafter.

\paragraph{Internal and External Tasks.}
The \emph{Sepsis} task splits patients from both hospitals, stratified by
hospital and whether they ever develop sepsis. The two external tasks
use the hospital-B patients in this test set as their shared test set.
\emph{Sepsis, new hospital} trains on all hospital-A patients.
\emph{Sepsis, 500 local labels} adds 500 labeled
hospital-B patients sampled from outside the test set, stratified by
whether they ever develop sepsis.

\FloatBarrier
\section{Data Access and Handling}
\label{app:data}
\paragraph{Data Sources and Access.}
Table~\ref{tab:app-datasets} lists the data sources and access conditions.
MIMIC-IV and eICU-CRD require credentialed access through PhysioNet \citep{pollard_physionet_2026};
SEER Research data require a
data-use agreement. The other sources are publicly accessible.
The National Center for Health Statistics is responsible only for the initial NHANES linked mortality data, not for the analyses, interpretations or conclusions in this paper.

\paragraph{Data Handling.}
Agents access training data and test features through code executed in
Apptainer containers. Test outcomes remain outside the agent environment
and are used for evaluation after each run. LLM interactions can include
data excerpts and summaries returned by the agents' code.

\FloatBarrier
\section{Agent Configurations}
\label{app:agents}
\begin{table*}[!htbp]
\centering
\caption{Agent systems, search strategies and submission selection.}
\label{tab:agents}
\footnotesize
\setlength{\tabcolsep}{6pt}
\begin{tabular}{@{}lll>{\raggedright\arraybackslash}p{0.38\textwidth}@{}}
\toprule
Agent & Type & Search strategy & Submission selection \\
\midrule
Codex & coding agent & agent-directed & final \texttt{submission.csv} \\
Autoresearch & coding agent & greedy search & final \texttt{submission.csv} \\
AIDE & research agent & greedy tree search & best self-reported validation score \\
AI Scientist v2 & research agent & best-first tree search & submission from the selected node in the latest stage with a saved submission \\
\bottomrule
\end{tabular}
\end{table*}

\paragraph{Implementations and Task Inputs.}
We use Codex CLI v0.149.1, AIDE v0.2.2 and AI Scientist v2
(commit \texttt{96bd516}). Codex reads the task description from its working
directory. Autoresearch also receives \texttt{program.md}, reproduced
below, and instructions to follow it. Neither Codex configuration receives
starter code.
AIDE receives the description through its task-description interface and
the task files in its data directory. AI Scientist v2 receives the full
description in both the Abstract and Experiments fields of its idea file,
with task files copied into each node's working directory.
We disable AIDE's final report and AI Scientist v2's writing and review
stages.

\paragraph{Search and Execution Settings.}
We enforce a 12-hour wall-clock limit for each experiment. AIDE's step limit is
100,000. AI Scientist v2 allocates time across its four experimental stages
in proportion to their original iteration quotas, $20{:}12{:}12{:}18$.
At the start of each main stage, it divides the remaining time among the
remaining stages in these proportions. Each stage allows 100,000 iterations.
AI Scientist v2 uses four workers, three initial drafts, a maximum debugging
depth of three, and three-seed evaluation of its selected node at stage
transitions. Both frameworks allow 3,600 seconds per node execution.

\paragraph{Framework Adaptations.}
We patch AIDE and AI Scientist v2 to record failed script executions as
failed nodes and treat metrics with incompatible types or optimization
directions as incomparable. If an AI Scientist v2 sub-stage has no usable
implementation, the framework can use the best implementation from the
most recent earlier sub-stage within the same main stage that has a usable
implementation.

\paragraph{Submission Selection.}
For Codex and Autoresearch, we evaluate \texttt{submission.csv} when
the run ends. For AIDE, we use its reported validation metric to select the
highest-ranked node that is not marked as faulty by the agent and has
an available prediction file.
For AI Scientist v2, we inspect sub-stages from latest to earliest.
In each, we use the saved best-node identifier if that node is not marked
as faulty; otherwise, we compare nodes not marked as faulty using the
agent's metric rule.
We select the first such node with its own saved submission, skipping
sub-stages whose selected node has none. If this procedure finds no
prediction file, we evaluate the file left in the working directory.
Selection does not use test scores.

\begin{table}[htbp]
\centering
\caption{Tokens per run (millions; mean $\pm$ SD over three repetitions).}
\label{tab:app-tokens}
\footnotesize
\begin{tabular}{@{}lcc@{}}
\toprule
Agent & Input & Output \\
\midrule
AIDE & 0.65 $\pm$ 0.05 & 0.28 $\pm$ 0.02 \\
Codex & 26.88 $\pm$ 1.86 & 0.17 $\pm$ 0.01 \\
Autoresearch & 51.44 $\pm$ 13.23 & 0.30 $\pm$ 0.08 \\
AI Scientist v2 & 4.29 $\pm$ 0.63 & 1.69 $\pm$ 0.24 \\
\bottomrule
\end{tabular}
\end{table}

\paragraph{Example Task Description.}
The following is an excerpt from the task description for UCI Diabetes.
\begingroup\small
\begin{verbatim}
# Task: 30-day readmission prediction (Diabetes, 130 US hospitals)

Predict whether a hospital encounter of a diabetic patient is followed by a
**readmission within 30 days**. One row per ENCOUNTER; columns:

- `encounter_id` — the unit you predict for
- `patient_nbr` — patient identifier, **train only**. Patients can have
  several encounters; the hidden test split is at PATIENT level. The test file
  has NO patient identifier and its rows are shuffled: every test row stands
  on its own.
- ~40 features (demographics, admission details, diagnoses, lab counts,
  medications). Categorical columns arrive ordinal-encoded; missing = NaN.
- `label` (train only): 1 = readmitted within 30 days (~11% of encounters).

## Files

| file | contents |
|---|---|
| `train.csv` | ~81k encounters WITH labels |
| `test_features.csv` | ~20k encounters, features only |
| `sample_submission.csv` | valid null: the training prevalence for every encounter |
| `validate.py` | format checker — `python validate.py submission.csv` |

## What to submit

`./submission.csv` with exactly: `encounter_id, PredictedProbability`
(probability of readmission within 30 days, in [0,1]), one row per test
encounter.

## Scoring (hidden labels, from your file alone)

- **`auroc`** — primary, higher better
\end{verbatim}
\endgroup

\FloatBarrier
\noindent\begin{minipage}{\linewidth}
\paragraph{Autoresearch Protocol.}
In the main experiments, all Autoresearch runs receive the following
instructions \citep{karpathy2026autoresearch}:
\begingroup\small
\begin{verbatim}
# Research program: greedy iterative improvement

Your objective is the task in `description.md`: maximize its primary metric
on the hidden test via `./submission.csv`.

## Phase 0 -- working baseline first

Write your own experiment script that goes end-to-end as simply as possible:
load the provided data, hold out an internal validation split from the
training data, fit a deliberately simple model, print your validation
score(s), and write a format-valid `./submission.csv`. Confirm
`python validate.py submission.csv` passes before any improvement work -- a
valid submission on disk is your safety net.

## Loop (repeat until the time budget is spent)

1. Propose ONE focused change (features, preprocessing, model class,
   hyperparameters -- one idea at a time).
2. Run the script; read the validation score.
3. Append one line to `journal.md`: what you changed, the numbers, and
   KEEP or REVERT.
4. KEEP the change only if the validation primary metric beat the best so
   far; otherwise revert it exactly.

## Hard rules

- Your experiment script(s) and `journal.md` are yours to edit; do not modify
  the provided data files, `validate.py`, or this file.
- Tune only against your internal validation split, never against the test
  features.
- Keep individual runs short enough to iterate many times; a long training
  needs a clear reason in `journal.md`.
- The final `./submission.csv` must come from your best-validated version
  (re-run it if you experimented after your best result).
- Everything happens inside this directory; no network access.
\end{verbatim}
\endgroup

\end{minipage}

\FloatBarrier
\section{Simple Baseline Models}
\label{app:references}
\paragraph{Training and Aggregation.}
The simple baseline models use the same training data and evaluation code as the agents' models.
We train each model with random seeds 0--4 and average the five separately
computed scores and calibration errors.
Model settings are fixed across tasks within each family.

\paragraph{Survival Analysis.}
We use the \texttt{pycox} Cox proportional-hazards model with two hidden
layers of 32 units, batch normalization, dropout 0.1 and no output bias.
Training uses Adam with learning rate $10^{-3}$, batch size 256 and
64 epochs on the full training set, without
early stopping. For SUPPORT, we standardize columns \texttt{x0} and
\texttt{x7}--\texttt{x13} using training-set statistics and retain
\texttt{x1}--\texttt{x6} as supplied. For the other survival tasks, we
one-hot encode nonnumeric columns using the training categories, impute
missing values with training medians, and standardize features.
We estimate the baseline hazard from the training data and evaluate
survival curves on the supplied time grid as right-continuous step functions.

\paragraph{Classification.}
We use \texttt{HistGradientBoostingClassifier} with 300 boosting iterations,
learning rate 0.1 and early stopping disabled. For UCI Diabetes, MIMIC-IV
and both eICU tasks, the model uses the supplied numeric features, excluding
the outcome, record identifiers and the UCI patient grouping identifier.
The estimator handles missing values. For PhysioNet 2012, we summarize
each measured variable by its mean, minimum, maximum and last value over
the observation window and use these summaries as predictors.

\paragraph{Sequential Prediction.}
We sort records by patient and ICU hour and forward-fill measurements
within each patient. Features include the current-hour measurements,
ICU hour and a hospital indicator; columns with no observed training
values are excluded. We use \texttt{HistGradientBoostingClassifier} with
the classification settings above and a maximum of 31 leaves per tree.
We reserve 15\% of training patients, stratified by whether sepsis occurs,
to select the alarm threshold. Among 40 quantiles from 0.80 to 0.999 of
the holdout probabilities, we choose the threshold with the highest utility.
The model trained on the remaining patients produces the test predictions.
We repeat this procedure for each task, including the 500 labeled
target-hospital patients where provided.

\FloatBarrier
\section{Per-Task Results and Invalid Runs}
\label{app:pertask}
\paragraph{Submission Validity.}
A run is valid if its selected submission has exactly the required columns and
test-set identifiers, with no duplicate identifiers. Probabilities must be finite and in
$[0,1]$, binary predictions must be in $\{0,1\}$, and survival curves must
be non-increasing, as applicable. Missing, unreadable or malformed files
make a run invalid. Masking-test results do not change validity or scores.

\paragraph{Score Aggregation.}
Penalized scores assign zero to invalid runs. For each agent and repetition,
we average normalized scores equally across tasks, then report the mean
and sample SD of the three averages.
Non-penalized scores first average valid runs within each task, then weight
tasks equally. Their SD is reported only when all runs are valid. Negative sequential-prediction
utility scores are clipped to zero in the reported primary-score tables.
Under each convention, we rank agents within each task by their mean
normalized score. Rank 1 is best, ties receive their average rank, and the
simple baseline model is not ranked. Table~\ref{tab:main} reports the mean and SD
of these ranks across tasks.

For each external-validation comparison, we divide the normalized score on the
external-validation task by the score on the corresponding task without the population
shift for the same agent and repetition index. We average ratios over pairs with two
valid runs and a positive score on the corresponding task without the population shift. Simple baseline ratios use the mean scores over five repetitions.

\begin{table}[htbp]
\centering
\caption{Primary scores by task for valid runs (mean $\pm$ SD). $\dagger$ Two valid runs; $\ddagger$ one valid run.}
\label{tab:app-raw}
\footnotesize
\begin{tabular}{@{}lrrrrr@{}}
\toprule
Task & \shortstack{Simple\\baseline\\models} & AIDE & Codex & Autoresearch & AI Scientist v2 \\
\midrule
SUPPORT & 0.622 & $0.639 \pm 0.010$ & $0.651 \pm 0.000$ & $0.656 \pm 0.005$ & $0.630 \pm 0.013$ \\
METABRIC & 0.642 & $0.658 \pm 0.004$ & $0.659 \pm 0.008$ & $0.676 \pm 0.007$ & $0.647 \pm 0.018^{\dagger}$ \\
Rotterdam & 0.685 & $0.677 \pm 0.022$ & $0.694 \pm 0.002$ & $0.692 \pm 0.004$ & $0.694 \pm 0.002$ \\
Rotterdam $\to$ GBSG & 0.686 & $0.653 \pm 0.028$ & $0.693 \pm 0.007$ & $0.686 \pm 0.021$ & $0.666 \pm 0.023$ \\
SEER & 0.865 & $0.749 \pm 0.216$ & $0.862 \pm 0.019$ & $0.870 \pm 0.003$ & $0.861 \pm 0.011^{\dagger}$ \\
SEER, later period & 0.862 & $0.865 \pm 0.014$ & $0.872 \pm 0.002$ & $0.868 \pm 0.011$ & $0.835^{\ddagger}$ \\
NHANES & 0.835 & $0.858 \pm 0.003$ & $0.863 \pm 0.007$ & $0.860 \pm 0.001$ & $0.852 \pm 0.004^{\dagger}$ \\
UCI Diabetes & 0.668 & $0.659 \pm 0.042$ & $0.682 \pm 0.011$ & $0.688 \pm 0.002$ & $0.677 \pm 0.009^{\dagger}$ \\
PhysioNet 2012 & 0.497 & $0.512 \pm 0.034$ & $0.532 \pm 0.022$ & $0.538 \pm 0.005$ & $0.524 \pm 0.010$ \\
MIMIC-IV & 0.884 & $0.893 \pm 0.000$ & $0.893 \pm 0.001$ & $0.892 \pm 0.000$ & $0.889 \pm 0.003$ \\
eICU & 0.871 & $0.877 \pm 0.001$ & $0.877 \pm 0.001$ & $0.874 \pm 0.007$ & $0.871 \pm 0.008^{\dagger}$ \\
eICU, new hospitals & 0.868 & $0.874 \pm 0.001$ & $0.875 \pm 0.000$ & $0.875 \pm 0.000$ & $0.852 \pm 0.033$ \\
Sepsis & 0.381 & $0.415 \pm 0.036$ & $0.446 \pm 0.017$ & $0.440 \pm 0.009$ & $0.181 \pm 0.256^{\dagger}$ \\
Sepsis, new hospital & 0.199 & $0.168 \pm 0.146$ & $0.231 \pm 0.021$ & $0.268 \pm 0.020$ & $0.228^{\ddagger}$ \\
Sepsis, 500 local labels & 0.248 & $0.217 \pm 0.102$ & $0.299 \pm 0.031$ & $0.253 \pm 0.056$ & $0.124^{\ddagger}$ \\
\bottomrule
\end{tabular}
\end{table}

\paragraph{Invalid Submissions.}
AI Scientist v2 produced all 12 invalid submissions: five had malformed survival columns,
five had incorrect patient-hour rows, one had incorrect patient identifiers, and one had
incorrect encounter identifiers.

\FloatBarrier
\section{Calibration and Subgroup Performance}
\label{app:behavior}
\paragraph{Expected Calibration Error.}
We compute ECE from the selected predictions in each valid run. For classification
and sequential prediction, we partition probabilities into ten equal-width
bins and compute
\[
\mathrm{ECE}=\sum_b \frac{n_b}{N}
  \left|\bar p_b-\bar y_b\right|,
\]
where $n_b$ is the number of observations in bin $b$, $N$ is the total number
of observations, and $\bar p_b$ and $\bar y_b$ are the bin's mean predicted probability
and event frequency. Empty bins contribute zero. For sequential prediction,
observations are patient-hours. We evaluate averaged ranks submitted as
probabilities without applying recalibration.

For survival analysis, we use the grid points nearest the 25th, 50th and
75th percentiles of test event times. At each time, we sort patients by
predicted survival into ten equal-size groups and compare mean predicted
survival with the within-group Kaplan--Meier estimate.
We weight absolute differences by group size and average the resulting
errors over the three times. Groups with fewer than five patients are
skipped.

For Table~\ref{tab:ece}, we first average valid runs within each task, then
weight tasks equally within each class or across all tasks.
All simple baseline calibration errors are averaged over the same five
repetitions per task as the simple baseline scores.

\paragraph{Subgroups.}
Classification subgroup scores use AUROC, including for PhysioNet 2012.
We divide age into test-set thirds and retain the four largest
race/ethnicity categories, grouping the remainder as \emph{Other}.
Subgroup metrics require at least 50 observations for AUROC and 80 for
survival concordance. Survival subgroup scores use Antolini concordance
without the adjustment used for the primary score.
Table~\ref{tab:app-subgroups} reports the mean gap between each valid run's
highest and lowest subgroup scores. Subgroup scores for the simple baseline models are averaged
over five repetitions. SUPPORT and METABRIC use generic covariate names and are
excluded; sequential-prediction tasks are also excluded.

\begin{table}[htbp]
\centering
\caption{ECE by task for valid runs (mean $\pm$ SD). $\dagger$ Two valid runs; $\ddagger$ one valid run.}
\label{tab:app-calibration}
\footnotesize
\begin{tabular}{@{}lrrrrr@{}}
\toprule
Task & \shortstack{Simple\\baseline\\models} & AIDE & Codex & Autoresearch & AI Scientist v2 \\
\midrule
SUPPORT & 0.032 & $0.049 \pm 0.027$ & $0.035 \pm 0.008$ & $0.044 \pm 0.013$ & $0.057 \pm 0.036$ \\
METABRIC & 0.054 & $0.048 \pm 0.003$ & $0.132 \pm 0.152$ & $0.081 \pm 0.033$ & $0.053 \pm 0.001^{\dagger}$ \\
Rotterdam & 0.042 & $0.059 \pm 0.045$ & $0.042 \pm 0.011$ & $0.055 \pm 0.027$ & $0.082 \pm 0.050$ \\
Rotterdam $\to$ GBSG & 0.039 & $0.112 \pm 0.096$ & $0.040 \pm 0.001$ & $0.160 \pm 0.170$ & $0.103 \pm 0.111$ \\
SEER & 0.006 & $0.027 \pm 0.040$ & $0.015 \pm 0.010$ & $0.016 \pm 0.011$ & $0.008 \pm 0.005^{\dagger}$ \\
SEER, later period & 0.007 & $0.008 \pm 0.005$ & $0.253 \pm 0.210$ & $0.005 \pm 0.001$ & $0.013^{\ddagger}$ \\
NHANES & 0.011 & $0.011 \pm 0.003$ & $0.016 \pm 0.010$ & $0.064 \pm 0.019$ & $0.022 \pm 0.000^{\dagger}$ \\
UCI Diabetes & 0.008 & $0.440 \pm 0.090$ & $0.133 \pm 0.221$ & $0.005 \pm 0.001$ & $0.079 \pm 0.106^{\dagger}$ \\
PhysioNet 2012 & 0.096 & $0.153 \pm 0.180$ & $0.015 \pm 0.012$ & $0.009 \pm 0.001$ & $0.051 \pm 0.030$ \\
MIMIC-IV & 0.020 & $0.270 \pm 0.214$ & $0.137 \pm 0.222$ & $0.137 \pm 0.222$ & $0.146 \pm 0.215$ \\
eICU & 0.013 & $0.269 \pm 0.224$ & $0.140 \pm 0.224$ & $0.271 \pm 0.222$ & $0.015 \pm 0.003^{\dagger}$ \\
eICU, new hospitals & 0.016 & $0.271 \pm 0.222$ & $0.141 \pm 0.223$ & $0.013 \pm 0.002$ & $0.219 \pm 0.192$ \\
Sepsis & 0.014 & $0.003 \pm 0.002$ & $0.004 \pm 0.005$ & $0.166 \pm 0.284$ & $0.215 \pm 0.075^{\dagger}$ \\
Sepsis, new hospital & 0.021 & $0.025 \pm 0.036$ & $0.023 \pm 0.028$ & $0.096 \pm 0.097$ & $0.002^{\ddagger}$ \\
Sepsis, 500 local labels & 0.020 & $0.171 \pm 0.142$ & $0.167 \pm 0.276$ & $0.009 \pm 0.004$ & $0.016^{\ddagger}$ \\
\bottomrule
\end{tabular}
\end{table}

\begin{table}[htbp]
\centering
\caption{Subgroup performance gaps across valid runs. ``Same lowest group'' is the share matching the simple baseline models' lowest-scoring subgroup.}
\label{tab:app-subgroups}
\footnotesize
\setlength{\tabcolsep}{4pt}
\begin{tabular}{@{}lllrrr@{}}
\toprule
Task & Grouping & \shortstack[l]{Simple baseline models'\\lowest group} & Same lowest group & Agent gap & \shortstack[r]{Simple\\baseline\\models' gap} \\
\midrule
Rotterdam & Age & Oldest & 92\% & 0.074 & 0.089 \\
Rotterdam $\to$ GBSG & Age & Oldest & 25\% & 0.019 & 0.033 \\
SEER & Age & Youngest & 73\% & 0.023 & 0.024 \\
SEER & Race/ethnicity & American Indian/Alaska Native & 73\% & 0.036 & 0.049 \\
SEER, later period & Age & Youngest & 80\% & 0.027 & 0.026 \\
SEER, later period & Race/ethnicity & American Indian/Alaska Native & 70\% & 0.035 & 0.048 \\
NHANES & Age & Youngest & 91\% & 0.158 & 0.222 \\
NHANES & Sex & Male & 100\% & 0.035 & 0.032 \\
NHANES & Race/ethnicity & Other & 100\% & 0.041 & 0.056 \\
UCI Diabetes & Age & Oldest & 100\% & 0.080 & 0.080 \\
UCI Diabetes & Sex & Male & 73\% & 0.006 & 0.008 \\
UCI Diabetes & Race/ethnicity & Caucasian & 9\% & 0.100 & 0.080 \\
PhysioNet 2012 & Age & Oldest & 100\% & 0.066 & 0.074 \\
PhysioNet 2012 & Sex & Female & 100\% & 0.012 & 0.021 \\
MIMIC-IV & Age & Oldest & 100\% & 0.084 & 0.078 \\
MIMIC-IV & Sex & Female & 100\% & 0.008 & 0.005 \\
eICU & Age & Oldest & 100\% & 0.065 & 0.065 \\
eICU & Sex & Female & 100\% & 0.009 & 0.006 \\
eICU, new hospitals & Age & Oldest & 100\% & 0.069 & 0.066 \\
eICU, new hospitals & Sex & Female & 100\% & 0.010 & 0.008 \\
\bottomrule
\end{tabular}
\end{table}

\FloatBarrier
\subsection{Effect of Evaluation Objectives}
\label{app:calib-control}

\paragraph{Setup.}
We compare modeling choices and predictions under three evaluation
objectives on UCI Diabetes, the binary 30-day readmission task.
Each group contains three repetitions of AIDE,
Codex, Autoresearch and AI Scientist v2, all using Qwen3.8 Flash
with a 12-hour budget. Group A reuses the main experiment's AUROC runs;
groups B and C are new runs with the same data, submission format and
task rules. The task descriptions specify:
\begin{itemize}
\item A: maximize AUROC.
\item B: minimize Brier score.
\item C: maximize AUROC and minimize ECE, using ten equal-width bins.
\end{itemize}
Group C provides no weights or rule for combining the two objectives.
For Autoresearch, we also change the protocol's instruction to
``maximize its primary metric'' to ``minimize its primary metric'' in B
and ``optimize its primary metrics'' in C.

\begin{figure}[tbp]
\centering
\includegraphics[width=0.65\textwidth]{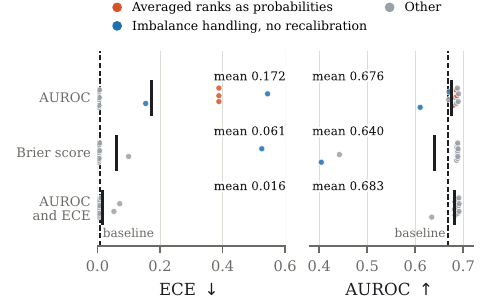}
\caption{AUROC and ECE under three evaluation objectives on UCI Diabetes. Points are valid runs; bars show means and dashed lines the simple baseline model. Colors mark the two modeling practices associated with large calibration error.}
\label{fig:calib-control}
\end{figure}

\paragraph{Evaluation.}
We evaluate the submitted probabilities using the same AUROC, Brier score
and ECE procedures as in the main experiment. The Brier score is
\[
\mathrm{Brier}=\frac{1}{N}\sum_{i=1}^{N}(p_i-y_i)^2,
\]
where $p_i$ is the predicted readmission probability and $y_i$ the binary
outcome. ECE uses the ten equal-width bins defined above.
Each group has 11 valid runs out of 12: one AI Scientist v2 run in each
group submitted 20,000 predictions for the 20,419 test admissions.
We pool all valid runs across the four agents and report means and sample
standard deviations.
Table~\ref{tab:calib-control-agents} reports results by agent, and
Figure~\ref{fig:calib-control} shows the AUROC and ECE distributions.

\begin{table}[!htbp]
\centering
\caption{Results by agent and evaluation objective on UCI Diabetes (mean $\pm$ SD over valid runs).}
\label{tab:calib-control-agents}
\footnotesize
\setlength{\tabcolsep}{4pt}
\begin{tabular*}{\textwidth}{@{\extracolsep{\fill}}llcccc@{}}
\toprule
Agent & Objective & Valid & AUROC & Brier score & ECE \\
\midrule
AIDE & AUROC & 3/3 & 0.659 $\pm$ 0.042 & 0.335 $\pm$ 0.068 & 0.440 $\pm$ 0.090 \\
AIDE & Brier score & 3/3 & 0.606 $\pm$ 0.142 & 0.099 $\pm$ 0.010 & 0.037 $\pm$ 0.054 \\
AIDE & AUROC and ECE & 3/3 & 0.685 $\pm$ 0.005 & 0.095 $\pm$ 0.002 & 0.022 $\pm$ 0.027 \\
\addlinespace
Codex & AUROC & 3/3 & 0.682 $\pm$ 0.011 & 0.161 $\pm$ 0.117 & 0.133 $\pm$ 0.221 \\
Codex & Brier score & 3/3 & 0.688 $\pm$ 0.001 & 0.093 $\pm$ 0.000 & 0.005 $\pm$ 0.002 \\
Codex & AUROC and ECE & 3/3 & 0.690 $\pm$ 0.002 & 0.093 $\pm$ 0.000 & 0.005 $\pm$ 0.001 \\
\addlinespace
Autoresearch & AUROC & 3/3 & 0.688 $\pm$ 0.002 & 0.093 $\pm$ 0.000 & 0.005 $\pm$ 0.001 \\
Autoresearch & Brier score & 3/3 & 0.688 $\pm$ 0.000 & 0.093 $\pm$ 0.000 & 0.004 $\pm$ 0.001 \\
Autoresearch & AUROC and ECE & 3/3 & 0.689 $\pm$ 0.002 & 0.093 $\pm$ 0.000 & 0.004 $\pm$ 0.001 \\
\addlinespace
AI Scientist v2 & AUROC & 2/3 & 0.677 $\pm$ 0.009 & 0.113 $\pm$ 0.026 & 0.079 $\pm$ 0.106 \\
AI Scientist v2 & Brier score & 2/3 & 0.545 $\pm$ 0.199 & 0.274 $\pm$ 0.255 & 0.265 $\pm$ 0.368 \\
AI Scientist v2 & AUROC and ECE & 2/3 & 0.658 $\pm$ 0.034 & 0.099 $\pm$ 0.007 & 0.039 $\pm$ 0.046 \\
\bottomrule
\end{tabular*}
\end{table}

\paragraph{Calibration and Discrimination.}
With AUROC and ECE as objectives, mean ECE falls from
$0.172\pm0.211$ in A to $0.016\pm0.023$ in C, while mean AUROC is
$0.683\pm0.016$, compared with $0.676\pm0.023$ in A.
The Brier group has mean ECE $0.061\pm0.157$ and AUROC
$0.640\pm0.107$. Across nine of its 11 valid runs, mean ECE is
$0.005\pm0.001$ and mean AUROC is $0.688\pm0.001$; the simple baseline model has
ECE 0.008 and AUROC 0.668. The other two runs contain modeling failures.
AIDE uses patient-level features derived from identifiers available only
in training and substitutes their means at test time, despite applying
temperature scaling (AUROC 0.442, ECE 0.099). AI Scientist v2 uses SMOTE
without recalibration (AUROC 0.404, ECE 0.525).

Two runs in C also retain appreciable calibration error after isotonic
calibration. AIDE maps its test predictions to the distribution of out-of-fold
predictions, then fits and evaluates isotonic regression on those same
out-of-fold predictions (test ECE 0.052). AI Scientist v2 includes the outcome label in an
aggregate feature used for training and validation, then fits
isotonic regression on predictions affected by this target leakage
(ECE 0.071).

\paragraph{Changes in Modeling Practice.}
We use Claude Opus 5.5 to annotate the code and execution records for all
22 valid runs in B and C without providing test scores, using the main
experiment's definitions of averaged ranks as probabilities, class-imbalance handling
and probability calibration. Seven runs are annotated again using the
same model; the repeated annotations agree on all recorded categories.
Group A uses the main experiment's annotations.

Table~\ref{tab:calib-control-practices} reports the resulting counts.
Agents use averaged ranks as probabilities in 3/11 runs in A and
none in B or C. Class weighting or oversampling without subsequent
recalibration occurs in 3/11, 1/11 and 0/11 runs, respectively.
Explicit calibration steps appear in 1/11 runs in A, 6/11 in B and
10/11 in C.
A seventh run in B uses logistic stacking, which the agent selects as an
ensemble rather than as a calibration step; it is excluded from the
explicit-calibration count.

\begin{table}[!htbp]
\centering
\caption{Modeling practices under the three UCI Diabetes objectives. Each objective has 11 valid runs; categories can overlap.}
\label{tab:calib-control-practices}
\footnotesize
\setlength{\tabcolsep}{4pt}
\begin{tabular*}{\textwidth}{@{\extracolsep{\fill}}>{\raggedright\arraybackslash}p{0.49\textwidth}ccc@{}}
\toprule
Practice & AUROC (A) & Brier score (B) & AUROC and ECE (C) \\
\midrule
Averaged ranks as probabilities & 3 & 0 & 0 \\
Class-imbalance handling without recalibration & 3 & 1 & 0 \\
Explicit probability calibration & 1 & 6 & 10 \\
\bottomrule
\end{tabular*}
\end{table}

\FloatBarrier
\section{Censoring and Real-Time Rule Violations}
\label{app:masking}
\begin{table}[!htbp]
\centering
\caption{Censoring errors in valid survival submissions; parentheses show the affected ensemble weight.}
\label{tab:app-censoring}
\footnotesize
\setlength{\tabcolsep}{5pt}
\begin{tabular}{@{}llc>{\raggedright\arraybackslash}p{0.50\textwidth}@{}}
\toprule
Agent & Task & Repetition & Problem \\
\midrule
Codex & SUPPORT & 2 & The labels of the XGBoost Cox member are all positive, so censored patients are treated as events (about 20\%) \\
Codex & SUPPORT & 3 & The labels of the XGBoost Cox member are all positive (about 20\%) \\
Codex & SEER, later period & 1 & The XGBoost Cox member adds 100,000 to censoring times, so censored patients stay in the risk set throughout (about 37\% on average, about half after 36 months) \\
Autoresearch & SEER & 1 & The labels of the XGBoost Cox member are not negated for censored patients (about 1.5\%) \\
Autoresearch & SEER & 3 & The binary members for 24, 48 and 60 months label patients censored before the time point as survivors (about 27\%) \\
Autoresearch & NHANES & 1 & The per-time-point IPCW classifiers take weights from a KM estimate that ignores the survey cycle, while censoring is determined entirely by the cycle; predicted survival in some cycles collapses to 0.1--0.3 (about 71\%) \\
Autoresearch & NHANES & 3 & As above; the per-time-point members carry more than half of the weight at later time points \\
Autoresearch & SEER, later period & 1 & IPCW weights only the cases and not the controls who survive past the time point, so death probabilities at later time points are systematically overestimated (about 60\%) \\
\bottomrule
\end{tabular}
\end{table}

\paragraph{Masking Procedure.}
To detect violations of the real-time rule, we remove later records from
each test patient's time series. For a patient with $n$ hourly records, we draw
$u\sim\mathrm{Uniform}(0.40,0.90)$ and retain the first
$\min(n,\max(4,\lfloor un\rfloor))$ records in time order.
One masked test file is reused across runs of the same task.
Training data and all other task files, including the sample prediction
file, remain unchanged.
We rerun the code producing the selected submission and compare alarms at
retained hours with those from the complete records.

We compute the fraction of changed alarms over all retained rows and over
the last retained row of each patient. A fraction above 0.1\% on either
measure flags a violation; zero on both is reported as
\emph{no change}. For the detected violations, rerunning the code on the
unmodified input changed no alarms.

Among the 13 valid AIDE and AI Scientist v2 runs tested, three AIDE runs violate the rule:
one on \emph{Sepsis, new hospital} and two on \emph{Sepsis, 500 local labels}.
One AI Scientist v2 rerun on the full input did not reproduce the selected predictions;
its masked and full reruns nevertheless agreed on all retained alarms.

\FloatBarrier
\section{Model Development}
\label{app:search}
We archive prediction files during model development and evaluate each version on the
test set after the run ends. A submission version is valid if its prediction file passes
the submission-format checks.

\begin{figure}[tbp]
\centering
\includegraphics[width=\textwidth]{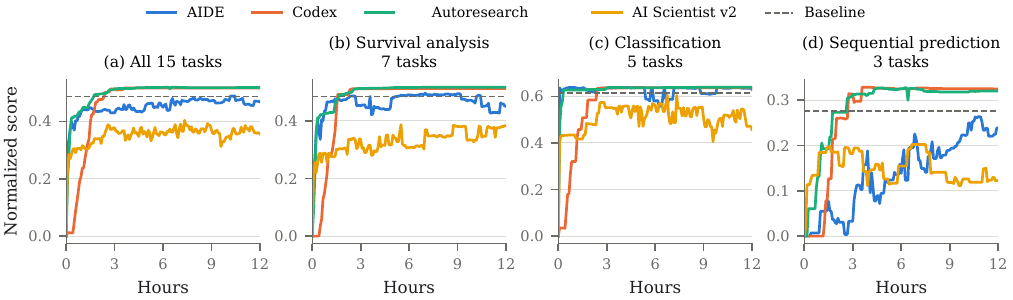}
\caption{Test performance over development time, averaged across runs; curves are zero before the first valid submission.}
\label{fig:search}
\end{figure}

\paragraph{Development Trajectories and Selection.}
Figure~\ref{fig:search} shows test performance over development time, and
Table~\ref{tab:development-summary} summarizes development pace and submission selection.
\begin{table}[!htbp]
\centering
\caption{Development and submission-selection summary.}
\label{tab:development-summary}
\footnotesize
\setlength{\tabcolsep}{8pt}
\begin{tabular}{@{}lcccc@{}}
\toprule
 & AIDE & Codex & Autoresearch & AI Scientist v2 \\
\midrule
Versions per run & 25.3 & 6.4 & 7.8 & 41.0 \\
Time to first valid version (min) & 30 & 64 & 17 & 30 \\
Run time (min) & 720 & 295 & 429 & 720 \\
Ended before 12 h & 0\% & 98\% & 100\% & 0\% \\
Selected submission has highest test score & 22\% & 32\% & 36\% & 21\% \\
Reported validation $-$ test score$^{\dagger}$ & $+0.047$ & $+0.006$ & $+0.008$ & $+0.012$ \\
\bottomrule
\end{tabular}
\par\smallskip
\footnotesize $^{\dagger}$AUROC and $C^{td}$ tasks only.
\end{table}

\paragraph{Version-Pair Analysis.}
Figure~\ref{fig:changes} compares valid submission versions from valid runs.
For AIDE and AI Scientist v2, we pair a version with the latest earlier
valid version from the nearest ancestor node that saved one.
For AI Scientist v2, some file-to-node and parent links are inferred.
For Codex and Autoresearch, pairs contain consecutive valid versions
in time; the earlier version need not be the basis for the later one.

Within each run, we order pairs chronologically. If there are more than
three, we divide them into three consecutive groups of approximately
equal size and sample one pair at random from each group.
With at most three pairs, we retain all and assign them to early, middle
and late in order. Group boundaries are based on pair counts.

Annotations compare the paired code and supporting execution records.
Figure~\ref{fig:changes} counts only changes recorded as affecting the
submitted predictions. Ensembling covers members, weights or averaging;
hyperparameter tuning covers settings within a model family; feature
engineering covers constructing, removing or selecting predictors.
Imputation, encoding and scaling are classified separately as preprocessing.
A pair can belong to multiple categories but counts once in each.
Each share is the fraction of sampled pairs with that change for the same
agent and group. Runs with fewer than three pairs contribute to fewer groups.

Table~\ref{tab:app-change} reports score changes in pairs containing each change type.
\begin{table}[!htbp]
\centering
\caption{Score changes for sampled version pairs containing each revision type, pooled across agents. Mean score change is later minus earlier normalized test score.}
\label{tab:app-change}
\footnotesize
\setlength{\tabcolsep}{5pt}
\begin{tabular}{@{}lccc@{}}
\toprule
& \multicolumn{2}{c}{Share of pairs} & \\
\cmidrule(lr){2-3}
Change & Score increase & Score decrease & Mean score change \\
\midrule
Ensembling & 58\% & 41\% & $+0.010$ \\
Hyperparameter tuning & 50\% & 49\% & $+0.004$ \\
Feature engineering & 51\% & 46\% & $+0.015$ \\
Model family & 79\% & 18\% & $+0.084$ \\
Bug fix & 66\% & 32\% & $+0.059$ \\
\bottomrule
\end{tabular}
\end{table}

\FloatBarrier
\section{Modeling Methods and Code Annotation}
\label{app:reading}
\paragraph{Code Annotation.}
We use Claude Opus 5.5 to annotate the code and execution records for the
selected submissions from all 180 runs. Method frequencies and modeling-practice
statistics use the 168 valid runs. For AIDE and AI Scientist v2, the inputs
include the selected node's script and execution output. For Codex and
Autoresearch, they include the scripts and commands that produced
the selected prediction file. Each annotation gives a categorical response
and cites supporting evidence by file and line number.

The codebooks cover model families, features, missing-data handling,
validation, tuning, ensembling and output processing, together with
modeling practices such as handling censoring and class imbalance.
We record practices used to produce the submitted predictions separately
from exploratory checks, such as evaluating demographic subgroups.
The practice codebook permits uncertain and not-applicable responses.
Changes during model development are annotated from sampled version pairs
(Appendix~\ref{app:search}).

Repeated annotation by the same model assesses consistency. It covers
10 runs for modeling methods and 36 for modeling practices
(Table~\ref{tab:app-reading}). The repeated practice annotations use the same
codebook, with instructions not to consult the first annotations.
Agreement was 124 of 130 method fields and 273 of 288 modeling-practice
fields. The modeling-practice comparison excludes free-text fields and treats
equivalent response labels as matches. Modeling-practice disagreements
that changed reported categories were checked against the cited code and
execution records. Final summaries include the recorded corrections.
Table~\ref{tab:app-reading} also compares these annotations with earlier
code annotations for missingness features and an earlier review of
validation designs.

\paragraph{Modeling Practices.}
Table~\ref{tab:clin-agent} reports each practice as a percentage of eligible
valid runs. Averaged ranks as probabilities and performance checks by age
or sex use all valid runs. Class-imbalance handling includes class weighting or
resampling and is evaluated on classification and sequential-prediction
tasks. Missingness features include missing-value indicators and counts; we
assess them on EHR tasks and NHANES.
Validation under population shift means choosing a model or threshold using
labeled target-hospital patients, or source data modified to resemble the
target population or split by hospital or ICU unit. This analysis covers
Rotterdam $\to$ GBSG and the two sepsis external-validation tasks.

\paragraph{Calibration Differences by Modeling Practice.}
Figure~\ref{fig:calib-share}(a) groups valid runs into those that submit
averaged ranks as probabilities, those that handle class imbalance without
recalibration and do not submit averaged ranks, and all remaining runs
(``Other experiments''). Figure~\ref{fig:calib-share}(b) compares classification
and sequential-prediction runs that handle class imbalance and do not submit
averaged ranks, with or without subsequent recalibration. Recalibration
includes isotonic regression, Platt scaling or prevalence correction after
handling class imbalance; it is annotated separately from
\emph{Probability calibration} in Table~\ref{tab:methods}.

For a valid run $r$ on task $t(r)$, define
$d_r=\mathrm{ECE}_r-\mathrm{ECE}^{\mathrm{ref}}_{t(r)}$,
where the simple baseline ECE is the mean over the five simple baseline repetitions.
Let $V$ contain all valid runs and $G\subseteq V$ a practice group.
The group's share of runs is $|G|/|V|$, and
its share of the total ECE difference is
$\sum_{r\in G}d_r/\sum_{r\in V}d_r$.
Negative differences are retained. Practice-group means and shares pool
runs across agents and tasks; Table~\ref{tab:ece} first averages within
each task.

Table~\ref{tab:methods} in the main text reports the frequencies of model families and validation methods.
\begin{table}[!htbp]
\centering
\caption{Modeling practices by agent (percentage of eligible valid runs).}
\label{tab:clin-agent}
\footnotesize
\setlength{\tabcolsep}{5pt}
\begin{tabular}{@{}lcccc@{}}
\toprule
Practice & AIDE & Codex & Autoresearch & AI Scientist v2 \\
\midrule
Averaged ranks as probabilities & 20\% & 18\% & 9\% & 6\% \\
Class-imbalance handling, no recalibration & 21\% & 4\% & 13\% & 35\% \\
Performance checked by age or sex & 0\% & 4\% & 9\% & 0\% \\
Validation under population shift & 33\% & 89\% & 89\% & 40\% \\
Missingness features & 81\% & 89\% & 89\% & 58\% \\
\bottomrule
\end{tabular}
\end{table}

\begin{table}[!htbp]
\centering
\caption{Agreement in repeated annotations and comparisons with earlier code reviews.}
\label{tab:app-reading}
\footnotesize
\setlength{\tabcolsep}{4pt}
\begin{tabular}{@{}ll@{}}
\toprule
Comparison & Agreement \\
\midrule
Modeling methods, repeated annotation & 95\% (130 fields, 10 runs) \\
Modeling practices, repeated annotation & 95\% (288 fields, 36 runs) \\
Missingness features vs.\ earlier records & 91\% (100 runs) \\
Validation under population shift vs.\ earlier checks & 95\% (22 runs) \\
\bottomrule
\end{tabular}
\end{table}

\FloatBarrier
\section{Other LLMs}
\label{app:pending}
We also evaluate Codex with gpt-6-luna and Gemini 3.8 Flash, using three
repetitions on each of the 15 tasks (45 runs per LLM). The Qwen3.8 Flash
results come from the main Codex evaluation.
Table~\ref{tab:pending-llm} reports normalized scores, ECE, submission
validity and token usage, using the scoring and ECE procedures in
Appendices~\ref{app:pertask} and~\ref{app:behavior}.

\begin{table}[!htbp]
\centering
\caption{Codex with three LLMs across 15 tasks. Values are mean $\pm$ SD over three repetitions; ECE is averaged across tasks and token counts are millions per run.}
\label{tab:pending-llm}
\footnotesize
\setlength{\tabcolsep}{4pt}
\begin{tabular}{@{}lccccc@{}}
\toprule
LLM & Normalized score $\uparrow$ & ECE $\downarrow$ & Valid runs & Input tokens & Output tokens \\
\midrule
Qwen3.8 Flash & 0.517 $\pm$ 0.003 & 0.086 & 100.0\% $\pm$ 0.0\% & 26.88 $\pm$ 1.86 & 0.17 $\pm$ 0.01 \\
gpt-6-luna & 0.517 $\pm$ 0.004 & 0.041 & 100.0\% $\pm$ 0.0\% & 2.77 $\pm$ 0.12 & 0.04 $\pm$ 0.00 \\
Gemini 3.8 Flash & 0.520 $\pm$ 0.006 & 0.029 & 100.0\% $\pm$ 0.0\% & 7.29 $\pm$ 0.24 & 0.05 $\pm$ 0.00 \\
\bottomrule
\end{tabular}
\end{table}

\end{document}